\documentclass{article}

\usepackage{amsfonts,amsmath}

\newtheorem{theorem}{Theorem}[section]
\newtheorem{lemma}{Lemma}[section]
\newtheorem{corollary}{Corollary}[section]

\newcommand{\qed}{\hfill$\Box$\\}

\newcommand{\proof}{~\\ \noindent {\bf Proof: }}

\newcommand{\tV}{\tilde{V}}
\newcommand{\tM}{\tilde{M}}

\newcommand{\1}{{\bf 1}}

\newcommand{\dd}{\,\text{\rm d}}
\newcommand{\be}{\begin{eqnarray}}
\newcommand{\ee}{\end{eqnarray}}

\newcommand{\beqa}{\begin{eqnarray*}}
\newcommand{\eeqa}{\end{eqnarray*}}

\newcommand{\eg}{{\em e.g., }}
\newcommand{\ie}{{\em i.e., }}

\def\Reals{{\bf {\sf R}}}

\def\P{{\sf P}}

\begin{document}

\title{A quantum diffusion network}

\author{George Kesidis\\ 
CSE and EE Depts\\
Pennsylvania State University\\
kesidis@engr.psu.edu\\
~\\
~\\
Technical Report No. CSE 09-012, Aug. 10, 2009}

\maketitle

\begin{abstract}
Wong's  diffusion network is a stochastic, zero-input Hopfield
network \cite{hopfield} with a Gibbs stationary distribution
over a bounded, connected continuum \cite{wong}. 
Previously, logarithmic thermal
annealing was demonstrated for the diffusion network 
\cite{kesidis95,holley} and digital
versions of it were studied and applied to imaging \cite{yin}.
Recently, ``quantum" annealed 
Markov chains have garnered significant
attention \cite{morita-nishimori,das-chakrabarti} because
of their improved performance over ``pure" thermal annealing.
In this note, a joint quantum and thermal  version
of Wong's diffusion network is described and  its
convergence properties 
are studied. Different choices for
``auxiliary" functions are discussed, including those of the kinetic
type previously associated with quantum annealing. 
\end{abstract}

\section{Introduction}

The optimization of a function $V(x)$, $x \epsilon D$, when the dimension
of the space $D$ is large and multiple local minima exist, is a computationally
difficult problem. A class of stochastic algorithms, known as simulated
annealing, has been developed for the case where $D$ is countable
\cite{holley}. For optimization over a bounded continuum,
a diffusion network was proposed in \cite{wong} and its thermal
annealing properties were established in \cite{kesidis95} after \cite{holley}.
In this note, we study a quantum version of this system that,
unlike thermal annealing,
modify the objective function $V$ in a nonlinear, nonuniform way.
Quantum annealing
proposals in the past include those
involving the Shr\"odinger operator with
potential
$V$ \cite{apolloni}, and those that add an
auxiliary function to $V$ 
that depends on $\nabla V$ (\eg the Ising spin glass
model with an external field \cite{das-chakrabarti}).
We consider here the latter type. 
Generally, the intuition behind the use of
an  auxiliary function is to initially perform a greater
breadth of search than under pure thermal annealing search.

\section{Specification of a Quantum Diffusion Machine}

Consider a time-inhomogeneous system described by 
\begin{eqnarray}
\dd u(t) &  = &  - \nabla [V(x(t)) - \Gamma(t) \tilde{V}(x(t))] \dd t 
+ \Sigma (T(t),x(t)) \dd W(t) \nonumber \\
x(t) &  = &  G(u(t)) \label{machine}
\end{eqnarray}
where the first equation is a stochastic differential equation of the
It\^{o} type,
\[ \begin{array}{l}
\nabla \; \mbox{is gradient with respect to the $x$ variables}, \\
u(t) \in \Reals^n ~~\mbox{where}~~ u(t)=(u^{1}(t),...,u^{n}(t)),\; n\geq 1,\\
x(t) \in (-1,1)^{n} , \; \;  x(t)=(x^{1}(t),...,x^{n}(t)), \\
\mbox{$W(t)$ is $n$-dimensional Brownian motion}, \\
G:\Reals^{n} \rightarrow (-1,1)^{n},\\
V,\tilde{V}:[-1,1]^{n} \rightarrow \Reals,\;\; V,\tilde{V} \in C^2,\\
M:=\sup_{x\in [-1,1]^{n}} V(x)-\inf_{x\in [-1,1]^{n}} V(x) < \infty, \\
\tilde{M}:=\sup_{x\in [-1,1]^{n}} \tilde{V}(x)-\inf_{x\in [-1,1]^{n}} 
\tilde{V} (x)< \infty, \\
\mbox{$T\geq 0$ the deterministic thermal/temperature process, and} \\
\mbox{$\Gamma\geq 0$ the deterministic quantum parameter process.} \\
\end{array} \]
$G$ is such that $x^{k}_t=g(u^{k}_t)$ where $g$ is a sigmoid threshold function
commonly found in neural networks:
\[ g(u^{k}(t)) = \tanh (u^{k}(t)/w) ~~\mbox{with}~~ w>0. \]

If $\tilde{V}\equiv 0$ and $\Sigma \equiv 0$ 
then the two relations in (\ref{machine}) 
describe a continuous-time Hopfield network with 
Lyapunov function $V$ and no external inputs (easily
realized as a ``neural" network when $V$ is quadratic). 
If $\tilde{V}\equiv 0$ and 
\[ \Sigma (T(t),x_{t}) = \mbox{diag} \left(
\sqrt{\frac{2T(t)}{f(x^{1}(t))}},...,
\sqrt{\frac{2T(t)}{f(x^{n}(t))}} \right) \]
where
\[ f(y) = g'(g^{-1} (y)) = \frac{1}{w} (1-y^{2} ) \]
and $T>0$ is constant, then the
stationary distribution of the $x$ process is Gibbs \cite{wong}:
\be
\mu (x) &:=& \tfrac{1}{Z} \exp (-V(x)/T)  \label{Gibbs}
\ee
where $Z$ is the partition (normalization) function. 
This is immediately seen by applying It\^{o}'s rule to (\ref{machine}),
after which the Fokker-Planck operator 
\cite{Karatzas-Shreve} governing the
distribution $p$ of the $x$ process is seen to be:
\begin{eqnarray}\label{FP}
L_{\Gamma}(p) & = & \mbox {div} [ A( T\nabla p
+ p\nabla( V-\Gamma \tilde{V}))]
\end{eqnarray}
where
\[ A(x) = \mbox{diag}(f(x^{1}),...,f(x^{n})). \]
That is, $L_0(\mu)\equiv 0$.
Furthermore, if $T(t) = T(0)/ \log_2 (2+t)$ 
(logarithmic thermal cooling), $T(0)>2M$, 
and the global extrema of $V$ are assumed 
in the interior $(0,1)^n$,
then  time-inhomogeneous process $x_t$ converges 
in probability to the (ground state)
set that globally minimizes the objective function $V$ \cite{holley}.

If fixed $T,\Gamma >0$, then the invariant distribution is clearly
\be
\mu_\Gamma (x) & := &  \tfrac{1}{Z_\Gamma} \exp(-(V(x) - \Gamma \tV(x))/T).
\label{Gibbs-like}
\ee
So, if $\Gamma = \mbox{o}(T)$,  $\mu_\Gamma$ is like 
a Gibbs distribution 
in the sense that it tends to indicate
the globally minimizing (ground) states of $V$ as 
$T\rightarrow \infty$.

\section{Quantum convergence to the Gibbs invariant}

In \cite{morita-nishimori} (and as explained in the recent
survey \cite{das-chakrabarti}), a quantum annealing process is considered.
They show that a faster-than-logarithmic quantum cooling schedule,
$\Gamma(t)\downarrow 0$ as $t\rightarrow\infty$,
can be used to establish convergence to the Gibbs invariant for 
fixed $T>0$, 
\ie not to the ground states. We now prove the analogous result
for the diffusion network, subject to a
more rapid cooling schedule,
by adapting the thermal convergence proof in
\cite{holley,kesidis95}. To this end, we show how the distribution $m_t$ of
$x_t$ ``tracks" the  distribution $\mu_{\Gamma(t)}$ (note that
this is obvious for all sufficiently large $t$
if $\Gamma$ reaches zero
in finite time). 
As the proof is a more substantive variation of \cite{holley} than for
pure thermal annealing of the diffusion network,
we give it in greater detail here than we did in \cite{kesidis95}.

We begin by defining 
\begin{eqnarray*}
z_t & :=  & \int_{(0,1)^n} \frac{m_t^2(x)}{\mu_{\Gamma(t)}(x)} \dd x ,
\end{eqnarray*}
where 
\beqa
\dot{m}_t & =&  L_{\Gamma(t)}m_t.
\eeqa
Let $\gamma_{\Gamma}$ be the gap between $0$ and the rest of the spectrum
of $L_{\Gamma}$ \cite{brand}:
\begin{eqnarray}\label{Dirichlet}
\gamma_\Gamma & = & \inf_\phi \frac{2T
\int (\nabla \phi )^{T} A (\nabla \phi) \mu_{\Gamma} \dd x}
{\int\int (\phi (x) -\phi (y))^{2} \mu_{\Gamma} (x) \mu_{\Gamma} (y) \dd y\dd x}
\end{eqnarray}
subject to the constraint that $\phi$ is not constant, where
integration is over $(0,1)^n$.
Equivalently,
\[  \gamma (\Gamma) = \inf_{\phi \not\equiv 0} T
\int (\nabla \phi )^{T} A (\nabla \phi) \mu \dd x
\]
subject to
$\int\phi^{2}\mu_{\Gamma}\dd x=1$ and $\int\phi\mu_{\Gamma}\dd x=0$.

\begin{theorem}\label{z-bound}
For {\em any} nonincreasing, differentiable
 quantum schedule $\Gamma$ with $\Gamma(\infty)=0$
and any constant temperature $T>0$: 
\beqa
\dot{z}_t 
& = & 
\mbox{O}\left(
\left(1+ \frac{\tM}{2T}\cdot\frac{\dot{\Gamma}(t)}{\gamma(\Gamma(t))}
\right)^{-1} 
\right).
\eeqa
\end{theorem}

\proof
Take 
\beqa
\phi_t= m_t/\mu_{\Gamma(t)}.
\eeqa
By direct differentiation, 
\beqa
\dot{z}_t 
& = & 2\int \phi_t \dot{m}_t \dd x
-\int\phi^2_t \dot{\mu}_t \dd x \\
& = & 2\int \phi_t L_t m_t \dd x
-T^{-1}\dot{\Gamma}(t)\int(\tV -<\tV>)\phi^2_t \mu_t \dd x \\
& \leq & 2\int \phi_t L_t (\phi_t \mu_t) \dd x
-T^{-1}\dot{\Gamma}(t) \tM z_t\\
& = & -2T \int(\nabla \phi_t)'A(\nabla\phi_t)\mu_t \dd x
-T^{-1}\dot{\Gamma}(t) \tM z_t,
\eeqa
where the last step is integration by parts using $A(0)=0=A(1)$.
Thus, by the previous expression for $\gamma(\Gamma(t))$ (noting
$\int (\phi_t-1)\mu_t\dd x=0$),
\beqa
\dot{z}_t  & \leq  &
\gamma(\Gamma(t)) \int(\phi_t-1)^2\mu_t \dd x
-T^{-1}\dot{\Gamma}(t) \tM z_t,\\
& = & \gamma(\Gamma(t)) (z_t-1) -T^{-1}\dot{\Gamma}(t)\tM z_t,\\
& =  & \left(-2\gamma(\Gamma(t))-T^{-1}\dot{\Gamma}(t)\tM \right)z_t 
~~\forall t\geq 0.
\eeqa
Integrating in time, we get an inequality of the form 
$z_t \leq \alpha_t + \int_0^t \beta_s z_s \dd s$
where $\alpha_t := z_0 + \int_0^t 2\gamma(\Gamma(s)) \dd s$ and
\beqa
\beta_t & :=&  -2\gamma(\Gamma(t)) - T^{-1}\dot{\Gamma}(t) \tM .
\eeqa
So by applying Gronwall's lemma and then multiplying by 
$1\equiv\exp(-\int_0^t\beta_r\dd r) /\exp(-\int_0^t\beta_r\dd r)$,
we get
\beqa
z_t & \leq & \alpha_t + \int_0^t \alpha_s\beta_s
\exp(\int_s^t \beta_r\dd r) \dd s\\
& = & 
\frac{\alpha_t\exp(-\int_0^t\beta_s\dd s)+ \int_0^t \alpha_s 
\dd \exp(-\int_0^s \beta_r\dd r)}{\exp(-\int_0^t\beta_r\dd r)}\\
& = & \frac{z_0+ \int_0^t 2\gamma(\Gamma(s))
\exp(-\int_0^s \beta_r\dd r)\dd s}{\exp(-\int_0^t\beta_r\dd r)}
\eeqa
where the last step is integration by parts (resulting
in term cancellation in the numerator) and the fact that
$\alpha_0=z_0$ and $\dot{\alpha}_t \equiv 2\gamma_t$.
Now note that as $t\rightarrow\infty$, $\gamma(\Gamma(t))
\rightarrow \gamma(\Gamma(\infty)):=\gamma(0) >0$
and  $\dot{\Gamma}(t)\rightarrow 0$,
and therefore $\beta_t \rightarrow -2\gamma(0)<0$.
Thus, the numerator and denominator of the previous
display both diverge as $t\rightarrow\infty$.
Applying L'H\^{o}pital's rule gives that
\beqa
\dot{z}_t & \leq & \frac{2\gamma(\Gamma(t))}{-\beta_t}
~\mbox{as}~t\rightarrow \infty.
\eeqa
\qed

\begin{lemma}\label{gamma-bound}
$\exists ~c>0$, which does not depend on $T$ or $\Gamma$, such that 
\beqa
\gamma(\Gamma) & \geq & cT\exp(-2 M^*(\Gamma)/T)
\eeqa
where 
\begin{eqnarray}\label{M-star}
M^*(\Gamma) & :=&  \sup(V-\Gamma\tV)-\inf(V-\Gamma\tV).
\end{eqnarray}
\end{lemma}
\proof
By (\ref{Dirichlet}), 
$\gamma(\Gamma) \geq  cT\inf\mu_{\Gamma}/(\sup\mu_{\Gamma})^2$ where
\beqa
c & := & \inf_\phi \frac{2 \int (\nabla \phi )^{T} A (\nabla \phi)  \dd x}
{\int \int (\phi (x) -\phi (y))^{2}  \dd y\dd x}.
\eeqa
So,  $\gamma(\Gamma) \geq$
\beqa
 cT Z_\Gamma \exp(-\sup (V-\Gamma\tV)/T)/
\exp(-2\inf(V-\Gamma\tV)/T)~=\\
cT\exp(-M^*(\Gamma)/T)\int\exp(-[(V- \Gamma\tV)-\inf(V-\Gamma\tV)]/T) \dd x.
\eeqa
\qed

Completing our adaptation of the arguments in \cite{holley,kesidis95}:

\begin{corollary}\label{converge-to-gibbs}
For {\em any} nonincreasing, differentiable
 quantum schedule $\Gamma$ with $\Gamma(\infty)=0$
and any constant temperature $T>0$, there is a constant $K<\infty$
such that for any $S\subset (0,1)^n$,
\beqa
\P(x_t\in S ) 
& \leq & K
\left(\int_S \mu_{\Gamma(t)}(x)\dd x\right)^{1/2}
~\forall t\geq 0.
\eeqa
\end{corollary}
\proof
Let 
\be
B(\Gamma,\dot{\Gamma},t)  & := & 
\left(1+ \frac{\tM}{2cT^2}\dot{\Gamma}(t)
\exp(2M^*(\Gamma(t))/T)
\right)^{-1} .  \label{B-def}
\ee
By the previous lemma and theorem,
\be
\lim_{t\rightarrow\infty}z_t & \leq &
\lim_{t\rightarrow\infty} B(\Gamma,\dot{\Gamma},t) 
~=~ 1.  \label{asym-bound}
\ee
Thus, by the continuity of $z_t$, 
there exists a positive constant $K< \infty$ such that
$z_t \leq K^2$ for all $t \geq 0$.
So, by the Cauchy-Schwarz inequality, 
\begin{eqnarray*}
\P(x_t\in S ) & =&  \int \1_S m_t\dd x
~ =~  \int \1_S \phi_t \mu_{\Gamma(t)}\dd x\\
& \leq & 
(\int \phi_t^2 \mu_{\Gamma(t)}\dd x)^{1/2} 
(\int \1_S^2 \mu_{\Gamma(t)}\dd x)^{1/2}  \\
& \leq & 
(\int z_t m_t\dd x)^{1/2} 
(\int_S \mu_{\Gamma(t)}\dd x)^{1/2}. 
\end{eqnarray*}
Substituting $z_t\leq K^2$ completes the proof.
\qed

Note that $K$ will 
depend on the parameter $z_0$.

\section{Global optimization of joint annealing}

To interpret this result, note that as $t\rightarrow\infty$,
$\mu_{\Gamma(t)}$ defined in (\ref{Gibbs-like}) is tending
to the Gibbs distribution (\ref{Gibbs}) for fixed $T>0$. Therefore, if 
$T>0$ is small  and $S$
does not include the ground states of $V$ (\eg
$S=\{x\in(0,1)^n ~|~V(x)\geq \theta+ \inf V\}$ 
for some sufficiently large $\theta>0$), then
$\P(x_t\in S)$ will be small.  To sharpen this statement,
consider {\em joint} quantum and thermal annealing.

\begin{theorem}
If $\Lambda(t):= \Gamma(t)/T(t) \rightarrow 0$, \ie $\Gamma = \mbox{o}(T)$,
and $D(t):=1/T(t)=\log_2(2+t)/T(0)$ with $T(0)>2M$, then 
$\exists K^*<\infty$ such that
\beqa
\P(x_t\in S ) & \leq &  K^* \left(\int_S \mu_t(x)\dd x\right)^{1/2}
~\forall t\geq 0,
\eeqa
where $\mu_t$ is given by (\ref{Gibbs}) with $T=T(t)$.
\end{theorem}
\proof Argue as for (\ref{asym-bound}) that
\beqa
z_t 
& = & \mbox{O}\left(
\left(1+ \frac{\tM\dot{\Lambda}(t)-M\dot{D}(t)}{2\gamma_t}
\right)^{-1} 
\right)
~\mbox{and}\\
\gamma_t & \geq & \frac{c}{D(t)}\exp(-2D(t)M^*(\Gamma(t))),
\eeqa
and so conclude as in the previous corollary, where the
condition $T(0)>2M=2M^*(0)$ figures in the resulting exponent of
$(1+t)$ after substituting for $D$.  \qed

So, if $S$ does not contain any of the ground states of $V$, then
$\lim_{t\rightarrow\infty} \P(x_t\in S ) \rightarrow 0$.

\section{Discussion: Choices for auxiliary function}\label{aux-sec}

\subsection{Homotopy methods}

In ``homotopy" based search \cite{homotopy}, the auxiliary function is
taken to be 
\beqa
\tV & :=  & V-V_0
\eeqa
where $\Gamma(0) \approx 1$ and
$V_0$ is unimodal  (only one local minimum which is, of course, 
its global minimum). Therefore, the ground states of 
$V-\Gamma(t)\tV$ are quickly found initially 
(\ie when $t>0$ is small so that $V-\Gamma(t)\tV\approx V_0$).
Ideally, $V_0$ is the best such function approximating
$V$ if suitable ``global" information about $V$ is available to 
determine it {\em a priori}; in this case, the initial ground states
(of $V-\Gamma(t)\tV$ for small $t>0$)
are close to those of objective function $V$.

\subsection{Contracting the objective function}

Suppose that the auxiliary function is simply
\beqa
\tV & := & V 
\eeqa
and that $\Gamma(0)<1$. In this case, the quantum diffusion network
is performing a kind of thermal annealing from temperature 
$T/(1-\Gamma(0))$  down to $T>0$. 
So, this choice of auxiliary function 
has the effect of linearly contracting the objective function $V$,
as in ``pure" thermal annealing, thereby facilitating a greater breadth
of search initially.

An example nonlinear contraction of the objective function $V$ is 
obtained by using the auxiliary function
\be\label{quantum-aux0}
\tV(x) & := &  -\varepsilon^{{\rm T}} \nabla^2 V(x) \varepsilon,
~~x\in (-1,1)^n,
\ee
where  $\varepsilon$ is a fixed $n$-vector.
Note that $\tV(x)>0$, respectively $\tV(x)<0$, 
 when $x$ is a local maximum, respectively minimum, of $V$.

The use of the auxiliary (\ref{quantum-aux0})
may not result in significant contraction of the objective
function (\ie from $V$ to $V-\Gamma\tV$)
in situations where the peaks or valleys of $V$ are very
deep. In a one dimensional ($n=1$)
setting, we can deal with this problem in the case
where there is a local extremum ($V'=0$)
between successive points at which $V''=0$ (no saddle
points in particular) by
augmenting this 
auxiliary using ``kinetic" components (\ie involving $V'$) which are
typically associated with ``quantum" annealing, \eg
\be\label{kinetic}
\tV & := &  -(\varepsilon^2 + |V'|^2) V''.
\ee
This example has a natural multidimensional form: 
$\tV :=  -\varepsilon^{{\rm T}} \nabla^2 V \varepsilon
 -(\nabla V)^{{\rm T}} \nabla^2 V \nabla V$.
In the case where it is advantageous to further contract
the objective $V$ at the points where $V''=0$ ($V$ and
$V-\tV$ for ``quantum" auxiliary $\tV$  of
(\ref{kinetic}) are equal at these points), one can
similarly propose to augment the auxiliary function
with $-(\varepsilon^2 + |V''|^2) V'''$, {\em etc.}

\bibliographystyle{plain}

\end{document}